%% file: main.tex
\pdfoutput=1

\documentclass[11pt]{article}

\usepackage[]{ACL2023}

\usepackage{times}
\usepackage{latexsym}

\usepackage[T1]{fontenc}

\usepackage[utf8]{inputenc}

\usepackage{microtype}

\usepackage{inconsolata}

\usepackage{microtype}
\usepackage{color}
\usepackage{multirow}
\usepackage{graphicx}
\usepackage{booktabs}
\usepackage{amsmath}
\usepackage{bbm}
\usepackage{float}
\usepackage{pifont}
\usepackage{microtype}
\usepackage{color}
\usepackage{multirow}
\usepackage{graphicx}
\usepackage{booktabs}
\usepackage{amsmath}
\usepackage{bbm}
\usepackage{float}
\usepackage{pifont}
\newcommand{\cmark}{\color{green}\text{\ding{51}}}
\newcommand{\xmark}{\color{red}\text{\ding{55}}}

\title{Uni-Encoder: A Fast and Accurate Response Selection Paradigm for Generation-Based Dialogue Systems}

\author{
Chiyu Song\thanks{\ \ Equal contribution.}\ \ \textsuperscript{1,2},
Hongliang He\footnotemark[1]\ \ \textsuperscript{1,2},
Haofei Yu\textsuperscript{3},
Pengfei Fang\textsuperscript{4,2},
Leyang Cui\textsuperscript{5},
Zhenzhong Lan\thanks{\ \ Corresponding author.}\ \ \textsuperscript{2}
\\
\textsuperscript{1}{Zhejiang University}, \textsuperscript{2}{School of Engineering, Westlake University} \\ \textsuperscript{3}{Language Technologies Institute, Carnegie Mellon University}, \textsuperscript{4}{Southeast University}, \textsuperscript{5}{Tencent AI Lab}\\
\{songchiyu, hehongliang, lanzhenzhong\}@westlake.edu.cn\\
haofeiy@cs.cmu.edu, fangpengfei@seu.edu.cn, leyangcui@tencent.com\\
}

\begin{document}
\maketitle
\begin{abstract}

Sample-and-rank is a key decoding strategy for modern generation-based dialogue systems. It helps achieve diverse and high-quality responses by selecting an answer from a small pool of generated candidates. The current state-of-the-art ranking methods mainly use an encoding paradigm called \textit{Cross-Encoder}, which separately encodes each context-candidate pair and ranks the candidates according to their fitness scores. However, \textit{Cross-Encoder} repeatedly encodes the same lengthy context for each candidate, resulting in high computational costs. \textit{Poly-Encoder} addresses the above problems by reducing the interaction between context and candidates, but with a price of performance drop. In this work, we develop a new paradigm called \textit{Uni-Encoder\footnote{\ \ Uni means one or unified here.}}, that keeps the full attention over each pair as in \textit{Cross-Encoder} while only encoding the context once, as in \textit{Poly-Encoder}. \textit{Uni-Encoder} encodes all the candidates with the context in one forward pass. We use the same positional embedding for all candidates to ensure they are treated equally and design a new attention mechanism to avoid confusion. Our \textit{Uni-Encoder} can simulate other ranking paradigms using different attention and response concatenation methods. Extensive experiments show that our proposed paradigm achieves new state-of-the-art results on four benchmark datasets with high computational efficiency. For instance, it improves $R_{10}@1$ by $2.9\%$ with an approximately $4 \times$ faster inference speed on the Ubuntu V2 dataset.

\end{abstract}

\input{intro}
\input{related_work}

\input{method_new}
\input{Experiment_new}

\input{discussion}

\input{conclusion}
\input{limitations}

\bibliography{acl2023, custom}
\bibliographystyle{acl_natbib}
\input{appendix}

\end{document}

%% file: intro.tex
\section{Introduction}
One of the major milestones of artificial intelligence is the ability to converse freely in natural language. Researchers in this field are working on building open-domain dialogue systems capable of handling a variety of topics. Depending on the implementation, these works can be categorized as \textbf{retrieval-based} \citep{lowe2015ubuntu, tao2019multi, yuan2019multi} or \textbf{generation-based} \citep{vinyals2015neural, serban2016building}. Retrieval-based systems carry out conversations by selecting an optimal response from a \textbf{large} candidate pool, which shows advantages in producing fluency and relevant response. However, retrieval-based systems may be limited by the capacity of the pre-defined candidate pool. Generation-based systems generate reasonable responses by a sequence-to-sequence model. Previous work shows that generation-based systems tend to give repetition or contradictory responses \citep{contra-1, contra-2}.

\begin{table}[t]
\Huge
\centering
\resizebox{\linewidth}{!}{%
\begin{tabular}{lccc}
\hline
Paradigm &
  \begin{tabular}[c]{@{}c@{}}Context-Response\\ Full Attention\end{tabular} &
  \begin{tabular}[c]{@{}c@{}}Avoidance of\\ Context \\ Recomputation\end{tabular} &
  Performance \\ \hline
\vspace{5mm}Bi-Encoder    & \xmark & \cmark & 80.6\%          \\
\vspace{5mm}Cross-Encoder & \cmark & \xmark & 82.8\%          \\
\vspace{5mm}Poly-Encoder  & \xmark & \cmark & 80.9\%          \\ \hline
Uni-Encoder (Ours)        & \cmark & \cmark & \textbf{85.9\%} \\ \hline
\end{tabular}%
}
\caption{\textit{Uni-Encoder} maintains the full attention between context and candidates while only encoding the lengthy context once. It is both fast and accurate compared with existing paradigms. Performance is the $R@1$ values evaluated on the Ubuntu Dialogue Corpus V2, and we refer to \citet{humeau2019poly} for the results of \textit{Bi-}, \textit{Cross-}, and \textit{Poly-Encoder}. The pre-trained BERT weights are all from \citet{devlin2018bert}.}
\label{tab:paradigm}
\end{table}

To combine the advantage of both methods, \citet{adiwardana2020towards} proposed a ``sample-and-rank'' method, which first samples a \textbf{small} pool of candidate responses from the generator and then re-ranks the candidates to get the best response by a ranker. Because a ranking model can view the whole responses while a pure generation method can only generate answers based on partial information, sample-and-rank method often performs better than the pure sample method. Under the sample-and-rank framework, researchers have greater freedom to explore different ranking methods \citep{zhang2019dialogpt, roller2020recipes, bao2021plato, thoppilan2022lamda}. They can encode candidates on-the-fly and encode them with the context. \textit{Cross-Encoder} \citep{urbanek2019learning} is one such paradigm. It jointly encodes the historical context with every candidate using full attention and ranks them according to the context-candidate matching scores. Despite its superior performance, \textit{Cross-Encoder} repeatedly encodes the context for each candidate. Since contexts are often much longer than responses, the computation is slow for practical use. \textit{Poly-Encoder} \citep{humeau2019poly, roller2020recipes} mitigates the above problem by reducing the full attention at every layer of Transformer \citep{vaswani2017attention} to global attention at the last layer. However, later work \citep{gu2020speaker,gu2021partner,han2021fine} confirms the importance of full attention and still uses \textit{Cross-Encoder} as the base building block for response selection.

One interesting research question is whether there is a way to realize full attention between each context-response pair without repeatedly encoding the same long context.
To answer the above question, we proposed a new paradigm called \textit{Uni-Encoder}, as presented in Table \ref{tab:paradigm}. In this new paradigm, all the candidates are concatenated with the context and jointly input to the same encoder in one forward pass. In the end, a softmax classifier is used to decide which candidate needs to be selected. If we concatenate candidates and context, we will get two problems. First, it is challenging to learn a good set of representations for candidates as they have different positional embeddings. Second, the averaging effect of the attention mechanism makes it difficult to distinguish various candidates. To address the above two problems, we propose two modifications to the traditional encoder networks.

First, we use the same set of positional embeddings for all candidates so that they are all treated equally because each is a possible continuation of the given context.

Second, we also design a novel attention mechanism for our new paradigm that only allows context-candidate attention and forbids the candidates to attend to each other directly.

Through changing these two designs, \textit{Uni-Encoder} can simulate the effects of any other paradigm (\textit{Cross-}, \textit{Bi-} or \textit{Poly-Encoder}) by changing how context and candidate attend to each other and how many candidates are processed in a single forward pass.

We evaluate our new paradigm on four benchmark datasets: PersonaChat \citep{zhang2018personalizing},  Ubuntu Dialogue Corpus V1 \citep{lowe2015ubuntu}, Ubuntu Dialogue Corpus V2 \citep{lowe2017training}, and Douban Conversation Corpus \citep{wu2016sequential}. Empirical results show that our method achieves state-of-the-art performance, jointly with high computational efficiency.
For instance, our \textit{Uni-Encoder} has an absolute $2.9\%$  $R@1$ improvement over the state-of-the-art \textit{Cross-Encoder} on the widely used Ubuntu Dialogue Corpus V2 dataset. It also has a lower computational cost than \textit{Cross-Encoder} and is approximately four times faster at inference time.

Our source code and model checkpoints will be released for reproducibility and future research\footnote{https://github.com/dll-wu/Uni-Encoder}.

%% file: related_work.tex
\section{Related Work} \label{background} 

Neural approaches for open-domain dialogue have seen significant recent progress. Due to this progress, generation-based dialogue systems have started outperforming retrieval-based methods \citep{roller2020recipes} as they can handle a wider variety of topics. \citet{adiwardana2020towards} show that sample-and-rank provides much more diverse and content-rich responses than beam-search. An additional ranking step allows responses to have full attention/view over themselves and the context, while pure generation methods only have left attention/view. This different view is why an additional ranking process is needed. In this study, we particularly focus on improving this ranking process. 

Because scoring candidates given a context is a classical problem in machine learning, numerous methods \citep{urbanek2019learning,reimers2019sentence,adiwardana2020towards} have been developed over the years. We will only discuss a few closely related works. Please refer to \citet{humeau2019poly} for a more detailed discussion. 

\textit{Bi-Encoder} \citep{reimers2019sentence} encodes the context and the candidate separately, then scores the relatedness between their representations. Due to its simplicity and efficiency, \textit{Bi-Encoder} often serves as a baseline method when a new dataset introduces \citep{lowe2015ubuntu, dinan2018wizard}. One significant advantage of the \textit{Bi-Encoder} is that its response representations can be pre-computed as they are context-independent. However, in modern generation-based dialogue systems, this advantage becomes a weakness. It is not necessary to pre-encode responses that are generated on-the-fly. And without context-response interaction, the ranking performance is severely weakened. \textit{Poly-Encoder} \citep{humeau2019poly} improves the accuracy of the \textit{Bi-Encoder} by adding a learned self-attention layer on top of the context and candidate features extracted from both encoders. Nevertheless, \textit{Cross-Encoder} is preferable to generation-based dialogues systems in practice due to its high effectiveness \citep{urbanek2019learning, humeau2019poly}. Instead of encoding each context and response pair separately, they encode them jointly using a full attention mechanism.

Recent improvements in response selection are mostly on \textit{Cross-Encoder}. For example, \citet{li2021small} adapt contrastive learning to \textit{Cross-Encoder} with a specially designed strategy and obtain a significant performance gain. \citet{lu2020improving} and \citet{gu2020speaker} add speaker change information to the inputs showing a large improvement in the response selection task. \citet{whang2020effective} and \citet{han2021fine} further post-train the encoder on domain-specific data and see additional improvements. To further utilize target data, \citet{xu2020learning} and \citet{whang2021response} investigate some additional self-supervised learning tasks. These tasks served as additional objectives jointly trained with the response selection task. Unlike all the above improvements, our improvement is on the encoder itself and can incorporate these additional tricks.

%% file: method_new.tex
\section{Methods}

This section elaborates on the problem formulation of dialogue response selection, compares different paradigms to model this task, and describes our implementation of \textit{Uni-Encoder}.

\subsection{Problem Formulation} \label{view}

{Re-ranking methods formulate the multi-turn response selection as a set of binary classification tasks.

In practice, given a dialogue context $C = \{u_{1}, u_{2}, ..., u_{N}\}$, where $u_k, k = 1, \ldots, N$ denotes a single utterance from either speaker, the response selection task is required to choose an optimal response from a candidate pool, denoted by $P = \{r_{1}, r_{2}, ..., r_{M}\}$. Every candidate $r_i$ is respectively paired with the context $C$, denoted as $f(C, r_i)$. The encoding function $f$ yields a representation that later undergoes non-linear transformations to predict a value of 1 for a proper match and 0 otherwise.
}

However, this binary classification view is not an efficient way of training the encoder because we need to encode the context $C$ once for each pair of context-response comparisons. Instead, \citet{humeau2019poly} leveraged in-batch
negative training and viewed this task as a multi-choice selection problem. This formulation optimizes, e.g., $softmax(f(C) \cdot f(r_1), ... ,f(C) \cdot f(r_M))$ by a ground truth label that is one-hot on the index of the sole positive candidate.

\begin{figure*}[th]

  \centering

  \includegraphics[width=0.86\textwidth]{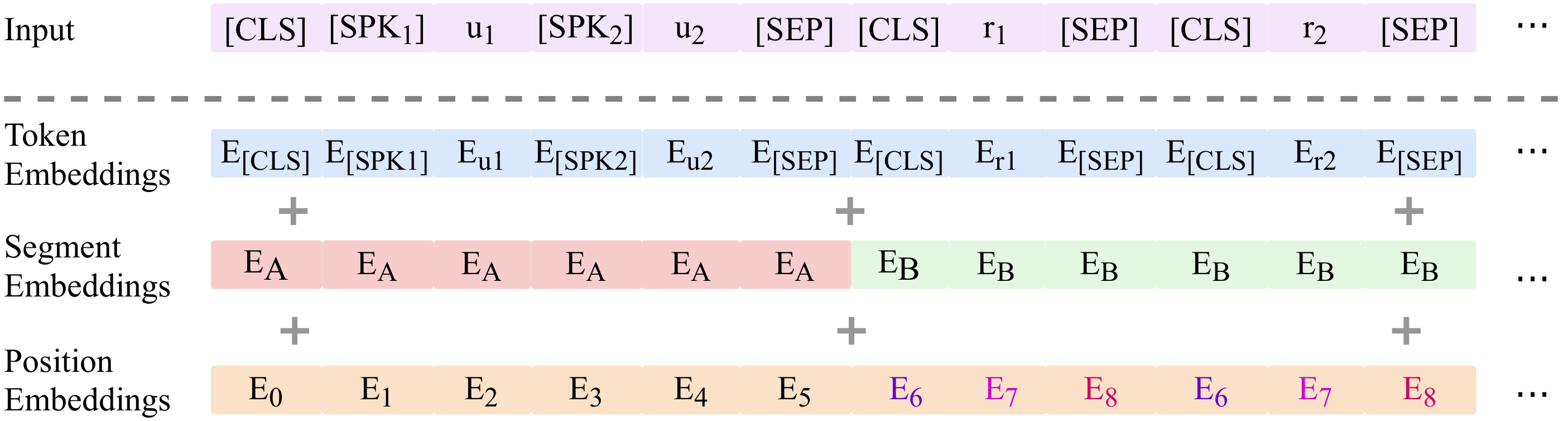}

  \caption{Input embeddings of the \textit{Uni-Encoder}. The positional embeddings of responses are repeated because each candidate is a possible continuation of the given context and should be treated equally. However, this new design will cause confusion among candidates. We address this problem by designing a new attention mechanism.}

  \label{fig:uni-input}

\end{figure*}

\subsection{Task Modeling Paradigms} \label{modeling}

In the following, we reuse the same set of notations in Section \ref{view}. Accordingly, \textit{Bi-}, \textit{Poly-}, \textit{Cross-}, and \textit{Uni-Encoder} model the response selection task as follows.

For \textit{Bi-Encoder}, selecting the proper response $r$ is picking the candidate that has the highest dot product with the context:

\begin{equation}\label{eq:r_bi}
f(C) \cdot f(r_1), ... ,f(C) \cdot f(r_M)
\end{equation}

where the response encoding is independent of the context encoding. \citet{humeau2019poly} show that, under the multi-choice view, the larger the $M$ is, the better the results are.

\textit{Poly-Encoder} is a variant of \textit{Bi-Encoder}. The only difference is that it adds an additional lightweight attention layer:

\begin{equation}\label{eq:r_poly}
\begin{aligned}
 g(f(C), f(r_1)), ... ,g(f(C),f(r_M))
\end{aligned}
\end{equation}
where $g$ is the light-weight attention component over the context and response representations generated by encoder $f$.

\textit{Cross-Encoder} has full attention between the context and responses. However, it has difficulty in taking the multi-choice view because it needs to re-compute the context for each candidate, which can result in a memory explosion. That is, for \textit{Cross-Encoder}, each context and response pair needs to go through the network $f$ together:

\begin{equation}\label{eq:r_cross}
f(C, r_1), ... ,f(C, r_M)
\end{equation}
In this way, for a batch containing $K$ context-response pairs, the heavy encoder $f$ needs to encode $K^2$ times, both computationally and memory intensive.

\textit{Uni-Encoder} also has full attention between the context and responses. Since all the candidate responses are concatenated and jointly encoded with the context in one forward pass, it naturally integrates the multi-choice view. Then the representation of each response is aggregated, and the most confident candidate is selected after feeding them into a softmax function:

\begin{equation}\label{eq:r_uni}
 softmax (f(C, r_1, ... ,r_M))
\end{equation}

Comparing formulas \ref{eq:r_bi} to \ref{eq:r_uni}, we can see that \textit{Bi-Encoder} has no interaction between context and responses in the encoding process; \textit{Poly-Encoder} allows partial interaction through a light-weight attention component; both \textit{Cross-} and \textit{Uni-Encoder} allow full interaction. Meanwhile, \textit{Uni-Encoder} avoids the drawback of \textit{Cross-Encoder} that repeatedly encodes the same lengthy context. Additionally, it establishes an exchange of information between candidates during the encoding process.

\subsection{Inputs to the Ranking Models: Same Positional Embedding for All Responses} \label{inputs_to_ranking}

We take the pre-trained BERT \citep{devlin2018bert} as our encoder. As illustrated in Fig. \ref{fig:uni-input}, the inputs to the BERT encoder consist of three components: the token embeddings, the segment embeddings help to distinguish between context and candidates, and the positional embeddings. In our setting, the positional embeddings for all the responses ($E_6$ to $E_8$ in Fig.\ref{fig:uni-input}) are repeated, treating each candidate as a coequal because they are all possible continuations of the context. We also have a separate speaker token for each utterance in the context to tell the model who is speaking. A \verb|[CLS]| and a \verb|[SEP]| token are placed before and after each candidate separately. 

\subsection{Attention Mechanisms: An Unified Ranking Framework}\label{attention_mechanism}
\begin{figure*}[th]

  \centering

  \includegraphics[width=\linewidth]{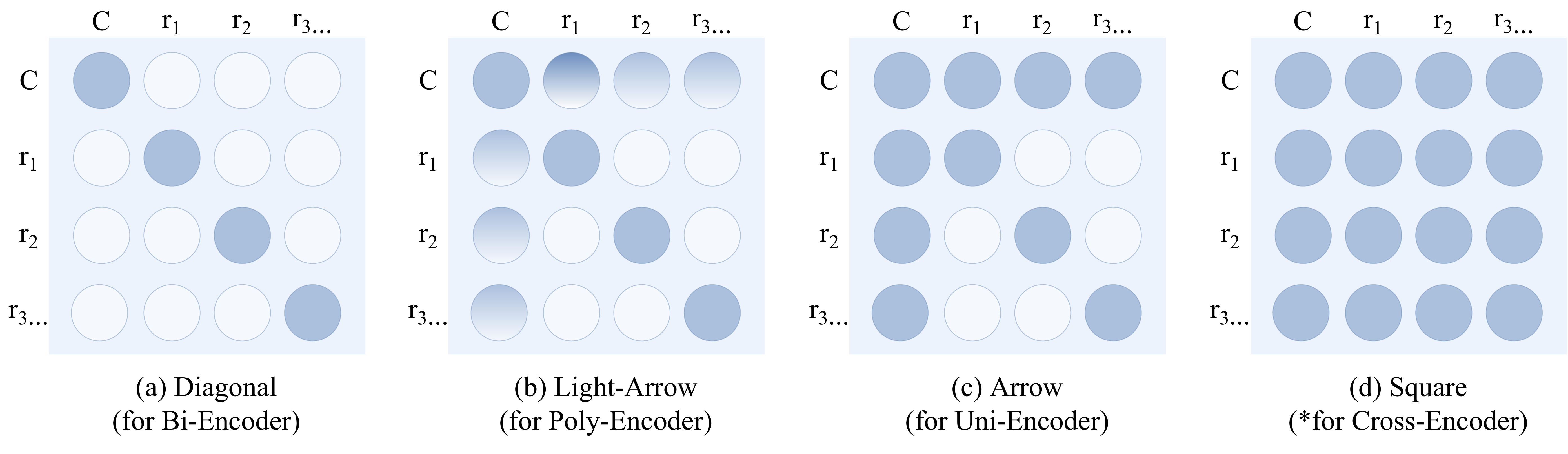}

\caption{The context-response attention maps corresponding to four paradigms, where attention is only allowed in filled areas \includegraphics[height=\fontcharht\font`\B]{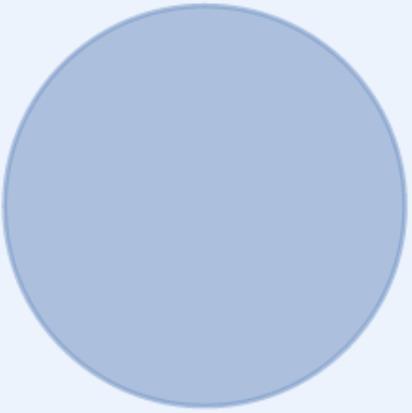}. The Arrow attention (c) is tailored for \textit{Uni-Encoder}, which realizes full attention between context and candidates and prevents candidates from directly attending to each other. The Light-Arrow attention (b) was introduced in \textit{Poly-Encoder} \citep{humeau2019poly}, where context and candidates only have attention in the last transformer layer. Changing the attention type and candidate number in parallel computation easily converts our work to other paradigms. For example, using the Diagonal attention (a) instead would make it a \textit{Bi-Encoder}, and *using the Square attention (d) while processing only one candidate at a time would make it a \textit{Cross-Encoder}.}

\label{fig:cam}
\end{figure*}

As Shown in Fig. \ref{fig:cam}, we design a new attention mechanism called Arrow Attention for \textit{Uni-Encoder}. Arrow Attention allows full attention between context and candidates while forbidding candidates from directly attending to each other. It realizes parallel processing of multiple candidates while only needing to process the context once. 

Fig. \ref{fig:cam} also shows that \textit{Uni-Encoder} can simulate other popular ranking frameworks by using different attention mechanisms. Specifically, (a) our work is equivalent to \textit{Bi-Encoder} if the Diagonal Attention is used instead, where the context and the candidates do not attend to each other. (b) The Light-Arrow Attention corresponds to \textit{Poly-Encoder}, where the context and candidates interact only at the last encoder layer through some additional light-weight attention. And the response representations are only available at the global feature level, e.g., the \verb|[CLS]| head or average token embedding. (c) The Arrow attention is tailored for \textit{Uni-Encoder}, where the context and the candidates have full attention, but the candidates do not attend to each other. (d) To test the extreme, we also have Square Attention, where all the context and responses attend to each other. However, it brings confusion among candidates as they share the same set of positional embeddings. The position confusion problem is addressed if it only processes one candidate at a time, which is equivalent to \textit{Cross-Encoder} by doing so.

%% file: Experiment_new.tex
\section{Experiments}

\subsection{Experimental Setup} \label{setup}

We initialize our implementation with the BERT \citep{devlin2018bert} checkpoint provided by the Huggingface package\footnote{https://huggingface.co/models}.  We also test post-training \citep{whang2021response, han2021fine} on top of pre-trained BERT when the checkpoints are available. The post-trained checkpoints are provided by \citet{han2021fine}. As introduced in Section \ref{background}, the post-training strategy is a common technique to adapt the general pre-trained knowledge to the target domain. In practice, it continues the models' pre-training on domain-specific texts before fine-tuning them on downstream tasks to attain better performances. All the experiments are run on six NVIDIA A100-SXM4-40GB GPUs with CUDA 11.1. We use the Noam scheduler and the Adam optimizer with $\beta_{1}=0.9$, $\beta_{2}=0.98$, and weight decay $=0.01$. For experiments on the Ubuntu Corpus V2, we use a peak lr of 2e-4. As we want each dataset to reach the maximum batch size in training, their learning rates are also adjusted accordingly in Section \ref{results}. As for the loss function, we add masked language modeling (MLM) loss on top of the classification loss with the same weight coefficients. We use the average token embedding from each candidate as the input to the softmax function. Models are all run until they converge, measured by a validation set.

\subsection{Dataset and Evaluation Metrics} \label{dataset}

In this section, we evaluate the proposed \textit{Uni-Encoder} across four standard datasets, i.e., PersonaChat~\citep{zhang2018personalizing},  Ubuntu Dialogue Corpus V1 \citep{lowe2015ubuntu}, Ubuntu Dialogue Corpus V2 \citep{lowe2017training}, and Douban Conversation Corpus \citep{wu2016sequential}.

\noindent\textbf{PersonaChat} \citep{zhang2018personalizing} is a crowd-sourced dataset with two-speaker talks conditioned on their given persona, containing short descriptions of characters they will imitate in the dialogue.

\noindent\textbf{Ubuntu Dialogue Corpus V1} \citep{lowe2015ubuntu} contains 1 million conversations about technical support for the Ubuntu system. We use the clean version proposed by \citet{xu2017incorporating}, which has numbers, URLs, and system paths replaced by special placeholders.

\noindent\textbf{Ubuntu Dialogue Corpus V2} \citep{lowe2017training} has several updates and bug fixes compared to V1. The major one is that the training, validation, and test sets are split into different periods. We choose this dataset to conduct a detailed study of \textit{Uni-Encoder} as it is the only dataset that \textit{Poly-Encoder} \citep{humeau2019poly} uses and has complete train/dev/test sets published.

\noindent\textbf{Douban Conversation Corpus} \citep{wu2016sequential} consists of web-crawled dyadic dialogs from a Chinese social networking website called Douban. Topics in this dataset are open-domain, and all the conversations are longer than two turns. Unlike other datasets where each context only has one proper response, the test set of Douban provides multiple proper responses.

\begin{table}[t]
\centering
\resizebox{0.96\linewidth}{!}{%
\begin{tabular}{@{}clccc@{}}
\toprule
\multicolumn{2}{c}{Dataset}                              & Train & Valid  & Test    \\ \midrule
\multirow{2}{*}{PersonaChat} & \multicolumn{1}{c}{Turns} & 65,719 & 7,801   & 7,512    \\
                             & Positive:Negative         & 1:19  & 1:19   & 1:19    \\ \midrule
\multirow{2}{*}{Ubuntu V1}   & \multicolumn{1}{c}{Pairs} & 1M    & 0.5M   & 0.5M    \\
                             & Positive:Negative         & 1:1   & 1:9    & 1:9     \\ \midrule
\multirow{2}{*}{Ubuntu V2}   & \multicolumn{1}{c}{Pairs} & 1M    & 195.6k & 189.2k  \\
                             & Positive:Negative         & 1:1   & 1:9    & 1:9     \\ \midrule
\multirow{2}{*}{Douban}      & \multicolumn{1}{c}{Pairs} & 1M    & 50k    & 6,670    \\
                             & Positive:Negative         & 1:1   & 1:1    & 1.2:8.8 \\ \bottomrule
\end{tabular}%
}
\caption{Statistics of four benchmark datasets.}
\label{tab:dataset}
\end{table}

\begin{table*}[th]
\small
\centering
{%
\begin{tabular}{@{}lcccccc@{}}
\toprule
\multirow{2}{*}{\begin{tabular}[c]{@{}l@{}}Paradigm\end{tabular}} &
  \multirow{2}{*}{Setup} &
  \multirow{2}{*}{\begin{tabular}[c]{@{}c@{}}Bs per GPU\end{tabular}} &
  \multicolumn{4}{c}{Ubuntu Corpus V2} \\ \cmidrule(l){4-7} 
                       &                                                                                   &   & $R_{10}@1$ & $R_{10}@2$ & $R_{10}@5$ & MRR   \\ \midrule
(i) \ \ \ Uni-Encoder  & \begin{tabular}[c]{@{}c@{}}Arrow Attn\\ w/ Res Concat\end{tabular}                   & 8 & \textbf{0.859} & \textbf{0.938} & 0.990 & \textbf{0.915} \\ 
\midrule
(ii) \begin{tabular}[l]{@{}l@{}}  \ \ Uni-Encoder \\ \ w/o Repeated Position ID  \end{tabular}     & \begin{tabular}[c]{@{}c@{}} Arrow Attn\\ w/ Res Concat \end{tabular} & 8 & 0.837 & 0.933 & \textbf{0.992} & 0.903 \\

\midrule
(iii) \ Concat-Cross-Encoder  & \begin{tabular}[c]{@{}c@{}}Square Attn\\ w/ Res Concat\end{tabular}                   & 8 & 0.826 & 0.916 & 0.980 & 0.892 \\ \midrule
(iv) \ \ Cross-Encoder    & \begin{tabular}[c]{@{}c@{}}Square Attn\\ w/o Res Concat\end{tabular}                  & 5 & 0.844 & 0.930 & 0.987 & 0.905 \\ \midrule
(v) \ \ \ Bi-Encoder        & \begin{tabular}[c]{@{}c@{}}Diagonal Attn\\ w/ Res Concat\end{tabular}                   & 8 & 0.835 & 0.925 & 0.987 & 0.899 \\ \midrule
(vi) \ \ \ Poly-Encoder       & \begin{tabular}[c]{@{}c@{}}Light-Arrow Attn (360)\\ w/ Res Concat\end{tabular} & 8 & 0.844 & 0.929 & 0.989 & 0.906 \\ 
\bottomrule
\end{tabular}%
}
\caption{Comparisons between different paradigms implemented according to the setups described in Section \ref{attention_mechanism}. By replacing the attention mechanism in \textit{Uni-Encoder}, a unified framework can simulate different paradigms, which optimally controls all other training variables for fair comparisons. Please note the \textit{Cross-Encoder} (iii) cannot reach the same large batch size as the others as it is more memory-intensive. For \textit{Poly-Encoder}, we choose the best setting with 360 context codes.}
\label{tab:attn}

\vspace{-1mm}

\end{table*}

The statistics of four benchmark datasets are shown in Table \ref{tab:dataset}. They vary greatly in volume, language, and topic. During training, we recycle the other labels in the same batch as negative samples instead of using the pre-defined negative candidates in each dataset. Several metrics are used to evaluate our model following previous works. We use $R_{c}@k$ to evaluate the model performance across four datasets. The mean reciprocal rank (MRR) metric is additionally calculated for PersonChat and Douban Conversation Corpus datasets. In the Douban Conversation Corpus, we also report the $P@1$ and mean average precision (MAP) values because it contains multiple positive candidates for a given context. It is also noted that the proportion of the positive and negative samples of the validation set is significantly different from that of the test set in the Douban Conversation Corpus. To alleviate this discrepancy, we also utilize the in-batch negative labels in the validation stage to determine an appropriate checkpoint for inference.

\begin{table*}[ht]
\centering
\resizebox{\textwidth}{!}{%
\begin{tabular}{@{}lcccccccccc@{}}
\toprule
\multirow{2}{*}{Models} &
  \multicolumn{3}{c}{Ubuntu Corpus V2} &
   &
   &
   &
  \multicolumn{2}{c}{PersonaChat} &
  \multicolumn{2}{c}{} \\ \cmidrule(lr){2-4} \cmidrule(lr){8-9}
 &
  $R_{10}@1$ &
  $R_{10}@2$ &
  $R_{10}@5$ &
   &
  \multicolumn{1}{l}{} &
   &
  $R_{20}@1$ &
  MRR &
  \multicolumn{1}{l}{} &
   \\ \midrule
BERT \citep{devlin2018bert} &
  0.781 &
  0.890 &
  0.980 &
   &
  \multicolumn{1}{l}{} &
   &
  0.707 &
  0.808 &
  \multicolumn{1}{l}{} &
   \\
Poly-Encoder 360 \citep{humeau2019poly} &
  0.809 &
  - &
  0.981 &
  \multicolumn{1}{l}{} &
  \multicolumn{1}{l}{} &
  \multicolumn{1}{l}{} &
  - &
  - &
  \multicolumn{1}{l}{} &
  \multicolumn{1}{l}{} \\
SA-BERT \citep{gu2020speaker} &
  0.830 &
  0.919 &
  0.985 &
   &
  \multicolumn{1}{l}{} &
   &
  - &
  - &
  \multicolumn{1}{l}{} &
   \\
BERT-CRA \citep{gu2021partner} &
  - &
  - &
  - &
   &
  \multicolumn{1}{l}{} &
   &
  0.843 &
  0.903 &
  \multicolumn{1}{l}{} &
   \\ \midrule
Uni-Encoder (Ours) &
  \textbf{\begin{tabular}[c]{@{}c@{}}0.859$^\star$\end{tabular}} &
  \textbf{\begin{tabular}[c]{@{}c@{}}0.938$^\star$\end{tabular}} &
  \textbf{\begin{tabular}[c]{@{}c@{}}0.990$^\star$\end{tabular}} &
   &
  \multicolumn{1}{l}{} &
   &
  \textbf{\begin{tabular}[c]{@{}c@{}}0.869$^\star$\end{tabular}} &
  \textbf{\begin{tabular}[c]{@{}c@{}}0.922$^\star$\end{tabular}} &
  \multicolumn{1}{l}{} &
   \\ \midrule
\multirow{2}{*}{} &
  \multicolumn{3}{c}{Ubuntu Corpus V1} &
   &
  \multicolumn{6}{c}{Douban Conversation Corpus} \\ \cmidrule(lr){2-4} \cmidrule(l){6-11} 
 &
  $R_{10}@1$ &
  $R_{10}@2$ &
  $R_{10}@5$ &
   &
  MAP &
  MRR &
  $P@1$ &
  $R_{10}@1$ &
  $R_{10}@2$ &
  $R_{10}@5$ \\ \midrule
BERT \citep{devlin2018bert} &
  0.808 &
  0.897 &
  0.975 &
   &
  0.591 &
  0.633 &
  0.454 &
  0.280 &
  0.470 &
  0.828 \\
SA-BERT \citep{gu2020speaker} &
  0.855 &
  0.928 &
  0.983 &
   &
  0.619 &
  0.659 &
  0.496 &
  0.313 &
  0.481 &
  0.847 \\
BERT-SL \citep{xu2020learning} &
  0.884 &
  0.946 &
  0.990 &
   &
  - &
  - &
  - &
  - &
  - &
  - \\
BERT+FGC \citep{li2021small} &
  0.829 &
  0.910 &
  0.980 &
   &
  0.614 &
  0.653 &
  0.495 &
  0.312 &
  0.495 &
  0.850 \\
UMS$_{BERT}$ \citep{whang2021response} &
  0.843 &
  0.920 &
  0.982 &
   &
  0.597 &
  0.639 &
  0.466 &
  0.285 &
  0.471 &
  0.829 \\
MDFN \citep{liu2021filling} &
  0.866 &
  0.932 &
  0.984 &
  \multicolumn{1}{l}{} &
  0.624 &
  0.663 &
  0.498 &
  0.325 &
  0.511 &
  0.855 \\
SA-BERT+HCL \citep{su2020dialogue} &
  0.867 &
  0.940 &
  0.992 &
  \multicolumn{1}{l}{} &
  0.639 &
  0.681 &
  0.514 &
  \textbf{0.330} &
  0.531 &
  0.858 \\
$^\clubsuit$UMS$_{BERT+}$ \citep{whang2021response} &
  0.875 &
  0.942 &
  0.988 &
  \multicolumn{1}{l}{} &
  0.625 &
  0.664 &
  0.499 &
  0.318 &
  0.482 &
  0.858 \\
$^\clubsuit$BERT-UMS+FGC \citep{li2021small} &
  0.886 &
  0.948 &
  0.990 &
  \multicolumn{1}{l}{} &
  0.627 &
  0.670 &
  0.500 &
  0.326 &
  0.512 &
  0.869 \\
$^\clubsuit$BERT-FP \citep{han2021fine} &
  0.911 &
  0.962 &
  \textbf{0.994} &
  \multicolumn{1}{l}{} &
  0.644 &
  0.680 &
  0.512 &
  0.324 &
  0.542 &
  \textbf{0.870} \\ \midrule
Uni-Encoder (Ours) &
  \begin{tabular}[c]{@{}c@{}}0.886\end{tabular} &
  \begin{tabular}[c]{@{}c@{}}0.946\end{tabular} &
  \begin{tabular}[c]{@{}c@{}}0.989\end{tabular} &
   &
  \begin{tabular}[c]{@{}c@{}}0.622\end{tabular} &
  \begin{tabular}[c]{@{}c@{}}0.662\end{tabular} &
  \begin{tabular}[c]{@{}c@{}}0.481\end{tabular} &
  \begin{tabular}[c]{@{}c@{}}0.303\end{tabular} &
  \begin{tabular}[c]{@{}c@{}}0.514\end{tabular} &
  \begin{tabular}[c]{@{}c@{}}0.852\end{tabular} \\
$^\clubsuit$Uni-Enc+BERT-FP (Ours) &
  \textbf{\begin{tabular}[c]{@{}c@{}}0.916$^\star$\end{tabular}} &
  \textbf{\begin{tabular}[c]{@{}c@{}}0.965$^\star$\end{tabular}} &
  \textbf{\begin{tabular}[c]{@{}c@{}}0.994\end{tabular}} &
  \multicolumn{1}{l}{} &
  \textbf{\begin{tabular}[c]{@{}c@{}}0.648$^\star$\end{tabular}} &
  \textbf{\begin{tabular}[c]{@{}c@{}}0.688$^\star$\end{tabular}} &
  \textbf{\begin{tabular}[c]{@{}c@{}}0.518\end{tabular}} &
  \begin{tabular}[c]{@{}c@{}}0.327\end{tabular} &
  \textbf{\begin{tabular}[c]{@{}c@{}}0.557\end{tabular}} &
  \begin{tabular}[c]{@{}c@{}}0.865\end{tabular} \\ \bottomrule
\end{tabular}%
}
\caption{Evaluations on four benchmark datasets. The models marked with $^\clubsuit$ have been post-trained, and the others are fine-tuned based on the naive BERT \citep{devlin2018bert}. $\star$ denotes statistical significance with p-value $<$ 0.05.}
\label{tab:full}
\end{table*}

\subsection{Validating Our Design Choices}
\label{attn}

In this section, we will validate our two design choices through a set of controlled experiments. As described in Section \ref{inputs_to_ranking} and \ref{attention_mechanism}, we are able to simulate different paradigms by replacing the attention mechanism in \textit{Uni-Encoder} with some minor modifications. We thus conduct experiments in this unified framework to control all other variables and make the fairest comparisons. Note that the \textit{Cross-Encoder} (iii) has to repeatedly encode the same lengthy context with every candidate, resulting in high memory usage and smaller batch size (5 in our experiments). The experimental results are shown in Table \ref{tab:attn}. 

\paragraph{Why Repeating Position ID for Responses?}
Let us first compare the results in Row (i) vs. Row (ii), where the only difference is that Row (i) use the same set of position IDs for all responses while Row (ii) has unique position IDs. Uni-Encoder with repeated position ID has significantly better results. This observation confirms our hypothesis that our responses should be treated equally.

\paragraph{Why using Full Attention Between Context and Responses?} If we compare the results of Row (i) with Row (v) and Row (vi), where the main differences lie in how much attention we have between context and responses, we can see that full attention can significantly boost performance. In fact, the more interaction (attention) they have, the better results they can get. Specifically, Poly-Encoder in Row(vi) has more interaction than Bi-Encoder in Row (v), and Uni-Encoder in Row (i) has more interaction than Poly-Encoder. These comparisons validate our design choices for full attention between context and responses.

\paragraph{Why Avoiding Attention Among Responses?}

Comparing results in Row (i) and Row (iii), we can see that if we allow attention among responses, the performance drops significantly. This is easy to understand because if we allow attention among responses, it will be difficult for the ranker to distinguish them. 

\paragraph{Why Avoiding Recomputing the Context?}
It is easy to understand that if we recompute the lengthy context, the computational time increases dramatically, which we will measure quantitatively in Section \ref{speed}. Here we show another dimension of the consequence of recomputing the context. As shown in Row (iv), the repetitive computation of the context stops the \textit{Cross-Encoder} from having a large batch size because of the memory constraint. However, a good enough batch size, hence negative samples, is important for a multi-choice setting, as examined in \citet{humeau2019poly}. As a result, the performance of \textit{Cross-Encoder} (iv) is only on par with \textit{Poly-Encoder} (vi).


\subsection{Comparison with State-of-the-Art Methods} \label{results}

We compare \textit{Uni-Encoder} with the existing state-of-the-art methods in Table \ref{tab:full}. Noted that, different from the comparison in Table \ref{tab:attn}, the methods in Table \ref{tab:full} are not entirely comparable as they have different additional training tricks. And these tricks often have a high impact on the performance of these methods. The only message we want to deliver here is that \textit{Uni-Encoder} can achieve state-of-the-art performance even without some of these complex training tricks.

For Ubuntu Corpus V1 and Douban Conversation Corpus, we also employ the advanced post-training model from \citet{han2021fine} and list the results separately with $^\clubsuit$ as it significantly affects the results and not all the methods use it.

As shown in Table \ref{tab:full}, \textit{Uni-Encoder} achieves the best overall performance across all four benchmarks. For example, it improves the $R@1$ value on PersonaChat, Ubuntu V1, and Ubuntu V2 datasets by $2.6\%$, $0.5\%$, and $2.9\%$, respectively.

However, \textit{Uni-Encoder} only achieves the best results on the Douban Corpus on four of the six metrics. We conjecture that the positive example size discrepancy between the training set and test set is the reason for its poorer performance. In \textit{Uni-Encoder}, we have chosen the multi-choice setting, assuming there is only one positive response. This setting allows us to leverage response concatenation and in-batch negative training to separate the positive sample from negative examples. However, multiple positive candidates in Douban Corpus at inference time (but not in training) break this assumption and may confuse the network. Our future study will quantify the impact of this assumption.

\textit{Uni-Encoder} also outperforms some of the more complex methods that rely on expensive training tricks, such as \citet{liu2021filling} adapted BiGRU to capture conversation-level representations, and \citet{su2020dialogue} leveraged hierarchical curriculum learning in their work. These approaches typically yield better outcomes, but at the expense of increased training budgets. In contrast, \textit{Uni-Encoder} only retains the MLM loss from pre-training and adds two extra tokens to distinguish between different speakers.

\begin{figure}[ht]

  \centering

  \includegraphics[width=\linewidth]{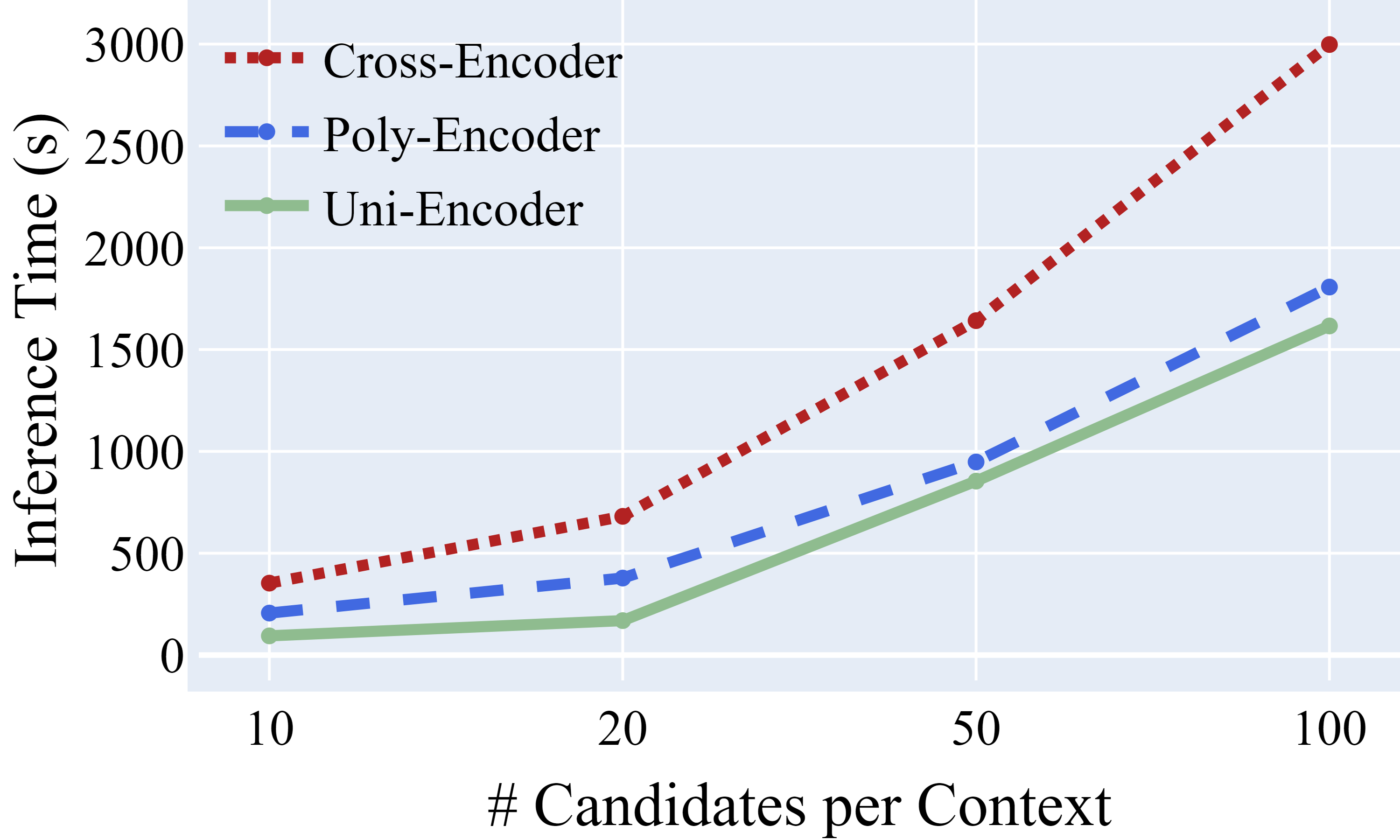}

\caption{The inference time comparison for \textit{Uni-Encoder} and other paradigms on the Ubuntu V2 test set. Please note that \textit{Poly-Encoder} cannot pre-compute candidate embeddings in a generation-based dialogue system, so the results differ from those reported in \citet{humeau2019poly} on retrieval tasks.}
\label{tab:speed}

\end{figure}

\subsection{Lower Computational Cost} \label{speed}
In addition to the accuracy gain, we also see that \textit{Uni-Encoder} is computational efficiency compared to other paradigms. We test it on the Ubuntu V2 test set (189,200 contexts). The implementation of \textit{Cross-} and \textit{Poly-Encoder} follows the method proposed in \citet{humeau2019poly}.

Despite the fact that candidate pools in generation-based dialogue systems are typically small, we are interested in understanding the performance of \textit{Uni-Encoder} with enlarged pools. To this end, we vary the pool size from 10 and 20 to 50 and 100 for each context by randomly selecting additional candidates from the corpus. We then conducted all speed tests on a single NVIDIA A100-SXM4-40GB with CUDA 11.1. The batch size for each paradigm was maximized as much as possible.
The results are presented in Figure 2. \textit{Uni-Encoder} demonstrates $4 \times$ faster inference speed compared to \textit{Cross-Encoder} when the pool size is appropriate. As the pool size increases, the advantages of \textit{Uni-Encoder} become more pronounced. Compared with \textit{Poly-Encoder}, \textit{Uni-Encoder} exhibits a similar trend, with slightly better overall efficiency. Furthermore, we have also deployed \textit{Uni-Encoder} in a commercial psychotherapy chatbot to rank the responses generated by large language models (LLMs). It has shown to be even more advantageous in this real-world dialogue application, as it returns results with only one forward pass, thus reducing the latency caused by other factors such as data transfer.

%% file: discussion.tex
\subsection{Qualitative Analysis}

To further understand the performance gap between different paradigms, we take the model checkpoints from Section \ref{attn} to go through examples that these methods predict differently. Some of the studied cases are shown in Table \ref{tab:examples} in Appendix. \textit{Uni-Encoder} is found to have the most specific and diverse selections. In contrast, even though some results of the other paradigms are not logically problematic, they sometimes prefer more generic responses. We conjecture this difference results from the fact that \textit{Uni-Encoder} compares and scores all the responses simultaneously. Candidates can still interact adequately with each other through their common attention to the context. With such an advantage, it would be easier to distinguish hard negatives from true positives.

%% file: conclusion.tex
\section{Discussion} \label{discussion}
This paper presents a new paradigm for the generation-based dialogue response selection task. Our proposed \textit{Uni-Encoder} avoids re-computing the lengthy context in the current state-of-the-art \textit{Cross-Encoder} method while maintaining the full context to candidate attention. Experimental results on four benchmark datasets show that our approach is both fast and accurate. As \textit{Uni-Encoder} holds the potential to build a more effective and efficient ranking paradigm, our future research will explore its usage in broader applications, such as improving the reward model in the reinforcement learning from human feedback (RLHF) framework \citep{stiennon2020learning, nakano2021webgpt, ouyang2022training}.

%% file: limitations.tex
\section{Limitations}

One major limitation of \textit{Uni-Encoder} is its suitability only for generation-based dialogue systems in which the number of responses is small. A two-stage approach is necessary for retrieval-based systems: Context-independent encoding methods like \textit{Poly-Encoder} first filter out a small set of candidates from the large pool, then \textit{Uni-Encoder} can pick out the best response from the pre-filtered collection. Moreover, as discussed in Section \ref{discussion}, \textit{Uni-Encoder} could be a good component of the RLHF approach. However, the increasing research of pure generation methods with alignments baked-in \citep{arora2022director, liu2023chain} may gradually replace the SFT+RL method. Consequently, \textit{Uni-Encoder} will have a smaller and smaller impact in terms of application. Nevertheless, because \textit{Uni-Encoder} unified all other ranking paradigms, we believe it remains helpful even as a theoretical framework.

%% file: appendix.tex
\appendix

\section{Qualitative Analysis}

\begin{table*}[t]
\small
\centering
\resizebox{0.96\linewidth}{!}{%
\begin{tabular}{@{}cll@{}}
\toprule
\# & \multicolumn{2}{c}{Examples}                                                                                 \\ \midrule
\multirow{4}{*}{1} &
  \multicolumn{2}{l}{A: have you looked in system settings \textgreater brightness and lock ? not power options} \\
 &
  \multicolumn{2}{l}{B: yes, of course. I'm here because the standard ways are failing on two my precise installations} \\ \cmidrule(l){2-3} 
   & \textbf{$^\star$Uni: care to post a screenshot?}                      & \textbf{Cross: I was just wondering} \\
   & \textbf{Bi: sry}                                                      & \textbf{Poly: Ah, ok.}               \\ \midrule
\multirow{5}{*}{2} &
  \multicolumn{2}{l}{A: Is there a way to force apt-get to install a package even if apt is locked by another running apt?} \\
   & \multicolumn{2}{l}{B: you don't want to do that wait till the updates are done then}                         \\
   & \multicolumn{2}{l}{A: It will take to long. Its a do-release-upgrade}                                        \\ \cmidrule(l){2-3} 
   & \textbf{$^\star$Uni/Cross: that will break things if you interupt it} &                                      \\
   & \textbf{Bi: Yes. I've done it several times}                          & \textbf{Poly: ok}                    \\ \midrule
\multirow{5}{*}{3} &
  \multicolumn{2}{l}{A: Does anyone know if there is a crossfeed plugin for Rhythmbox in the repositories?} \\
   & \multicolumn{2}{l}{B: why do want to feed rhythmbox?}                                                        \\
 &
  \multicolumn{2}{l}{A: crossfeed is a type of signal processing that removes the separation inherent in stereo recordings it's for headphone listening} \\ \cmidrule(l){2-3} 
   & \textbf{$^\star$Uni/Cross/Poly: it's called crossfade ;)}            &                                      \\
   & \textbf{Bi: could you explain more about what you want?}              & \textbf{}                            \\ \bottomrule
\end{tabular}%
}
\caption{Cases studied from Ubuntu V2 for comparing selections of different paradigms where $\star$ denotes the correct choice.}
\label{tab:examples}
\end{table*}